# HistoART: Histopathology Artifact Detection and Reporting Tool

Seyed Kahaki[†*], Alexander R. Webber[†], Ghada Zamzmi[†], Adarsh Subbaswamy, Rucha Deshpande, Aldo Badano

Division of Imaging, Diagnostics, and Software Reliability, Office of Science and Engineering Laboratories, Center for Devices and Radiological Health, U.S. Food and Drug Administration (FDA), MD

[†]Co-first Authors

**Abstract**

In modern cancer diagnostics, Whole Slide Imaging (WSI) is widely used to digitize tissue specimens for detailed, high-resolution examination; however, other diagnostic approaches, such as liquid biopsy and molecular testing, are also utilized based on the cancer type and clinical context. Whole Slide Imaging has revolutionized digital histopathology by enabling automated and precise analysis. However, WSIs are vulnerable to various artifacts introduced during slide preparation and scanning, which can compromise the reliability of downstream image analysis tasks. To address this challenge, we propose and compare three robust approaches for artifact detection and reporting in WSIs: (1) a foundation model-based approach (FMA) utilizing a fine-tuned Unified Neural Image (UNI) architecture, (2) a deep learning approach (DLA) built on a ResNet50 backbone, and (3) a knowledge-based approach (KBA) leveraging handcrafted features derived from texture, color, and frequency-based metrics. These methodologies target six prevalent artifact types in WSIs: tissue folds, out-of-focus regions, air bubbles, tissue damage, marker traces, and blood contamination. The approaches were evaluated on a dataset of over 50,000 image patches sourced from diverse WSI scanners, including Hamamatsu, Philips, and Leica Aperio AT2, across multiple imaging sites. Our foundation model achieved superior patch-wise AUROC performance of 0.995 (95% CI [0.994, 0.995]), outperforming the ResNet50-based method (AUROC: 0.977, 95% CI [0.977, 0.978]) and the knowledge-based approach (AUROC: 0.940, 95% CI [0.933, 0.946]). To bridge the gap between detection and actionable conclusions, we developed a quality report scorecard that quantifies the number of high-quality image patches in the dataset and visualizes the distribution of artifact subgroups. This comprehensive pipeline not only may enhance the reliability of WSI analysis but may also provide a scalable and interpretable framework for improving digital pathology workflows.

## I. Introduction

Digital histopathology, empowered by whole-slide imaging (WSI), has emerged as a cornerstone of modern medical diagnostics [1]. WSIs provide high-resolution digitized representations of histological slides, facilitating automated analysis and computational pathology (CPATH) applications. However, the reliability of WSI-based analyses are often undermined by artifacts introduced during tissue preparation, mounting, or scanning. These artifacts—such as out-of-focus blur, tissue folds, air bubbles, and markings—not only obscure diagnostic features but also introduce noise into image data, posing significant challenges for machine learning models and automated systems increasingly deployed in cancer diagnosis and other critical applications [2]. Artifacts arise from diverse sources: improper focus or misalignment during slide scanning can cause blurring; mounting issues, such as tissue folding or the presence of air bubbles, can distort cellular structures; and biopsy procedures may result in tissue or blood damage [2, 3]. For example, tissue folds, caused by misplacement during mounting, can obscure cellular details, while air bubbles disrupt imaging continuity. These artifacts compromise the integrity of downstream tasks like segmentation, classification, and biomarker quantification. Given that histopathology remains the gold standard for disease diagnosis, addressing these artifacts is necessary to ensure accurate and reliable analysis. The challenge is compounded by the immense size and resolution of WSIs, often comprising trillions of pixels, making artifact detection a non-trivial computational task.

In this paper, we propose methods for artifact detection in WSIs based on three complementary approaches: (1) a foundation model-based approach (FMA) leveraging a fine-tuned Unified Neural Image (UNI) model [4], (2) a deep learning-based approach (DLA) utilizing a fine-tuned ResNet50 architecture [5], and (3) a knowledge-based approach (KBA) incorporating handcrafted features based on texture, color, and frequency-based metrics. This multi-pronged approach enhances the artifact detection accuracy and robustness by combining the strengths of each method. The foundation model leverages pretraining on a large histopathology image dataset annotated for artifacts, such as out-of-focus blur, tissue folds, air bubbles, markings, and artifact-free regions. Fine-tuning this model enables it to effectively identify complex artifact patterns across diverse WSIs. To complement the foundation model, a ResNet50 architecture was trained on a similarly annotated dataset to bolster the pipeline's artifact identification capabilities. We use ResNet as it has been extensively used as the backbone and baseline model for a variety of downstream tasks [5]. Additionally, handcrafted feature extraction methods were integrated to quantify texture, color, and frequency-based metrics indicative of artifact presence. These features provide interpretability and serve as a complementary layer of analysis, capturing artifact patterns that may not be as easily discerned by deep learning and foundation methods alone. Figure 1 illustrates the proposed computational pathology (CPATH) pipeline. The preprocessing step uses the artifact detection models to filter out patches containing artifacts, ensuring that only artifact-free regions are passed to downstream diagnostic systems. This approach can mitigate the impact of artifacts on CPATH workflows [2]. Together, the proposed pipeline represents a comprehensive solution for artifact detection in WSIs.

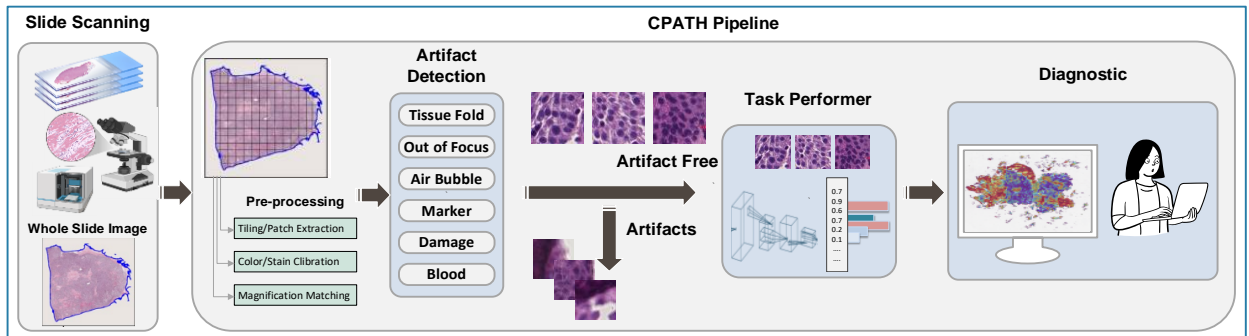

**Figure 1.** A Computational Pathology (CPATH) pipeline with a preprocessing pipeline designed to address artifacts in Whole Slide Images (WSIs). Artifact detection models can filter out patches with artifacts, ensuring that only artifact free regions are passed to the diagnostic for accurate predictions when necessary.

Artifact detection in the computational pathology (CPATH) literature often receives less attention than preprocessing methods, which are designed to reduce color variations and apply image augmentations [6]. While preprocessing strategies, such as stain normalization and augmentation, are essential for standardizing input data, they do not address the critical challenge of detecting and mitigating artifacts prior to diagnostic analysis. The detection of artifacts remains an underrepresented aspect of WSI preprocessing [7], despite its potential to significantly enhance the reliability of downstream tasks such as segmentation and classification. Some studies [8-10] have adopted quality control (QC) methods at lower magnifications, focusing on rapid identification of faulty WSIs. For instance, Avanaki et al. (2021) [11] proposed a quality estimation framework that combines metrics such as blurriness, contrast, and brightness to determine whether a WSI is accepted or discarded based on a reference threshold. However, such methods typically focus on global image quality and lack the granularity to identify specific artifacts. Similarly, Bahlmann et al. (2020) [10] employed texture-based features and stain absorption metrics to differentiate diagnostically relevant regions from irrelevant ones. While effective in certain scenarios, this approach is prone to missing artifacts embedded within diagnostically relevant areas, particularly at lower magnifications. Further, their validation was constrained to specific staining protocols and tissue types, highlighting the need for methods that generalize across diverse imaging conditions and artifact types. Haghighat et al. [12] developed PathProfiler, a patch-level classification algorithm based on ResNet18, for analyzing WSIs in coarse regions. This method, designed for a single histological domain (prostate specimens), uses an innovative scoring system to assess slide usability by evaluating parameters like staining quality and the presence of out-of-focus and folding artifacts. While open source, its application is largely limited to prostate specimens and faces generalization challenges, as shown in subsequent validation studies. Additionally, it does not account for other common WSI artifacts, such as blood, marker, or air bubbles, which can impact usability. The semi-automatic HistoQC [13] is one of the most widely used tools for analyzing domain shifts and quality variations in WSIs. It effectively assesses staining intensity, color variability, and other cohort-level shifts. However, its utility for artifact detection is limited by its focus on slide-level QC rather than pixel-level analysis. HistoQC's validation primarily involved comparing its performance to human analysts in categorizing WSIs as qualified or disqualified, with a specific focus on kidney biopsies. Recently, Kanwal et al. (2024) [2] introduced a probabilistic model that integrates a convolutional neural network (CNN) feature extractor with a sparse Gaussian Processes (GPs) classifier. This approach enhances the performance of state-of-the-art deep convolutional neural networks (DCNNs) for artifact detection and offers robust uncertainty estimates. Their model achieved AUC scores of 0.99 for blur detection and 0.93 for folded tissue detection on previously unseen data. In our study, we compared our results with those reported by Kanwal et al. [2] and present the findings in the results section.

Artifact detection in WSIs is inherently challenging due to the diverse nature of artifacts and their varying impacts on diagnostic reliability. Artifacts such as blurriness can often be addressed through rescanning or deblurring techniques, while others, like tissue folds or air bubbles, necessitate robust detection systems to ensure diagnostic reliability. Automatic identification of artifact-free regions has been shown to enhance dataset quality and improve the performance of downstream tasks, including segmentation and classification [14]. Moreover, preprocessing steps that address artifacts strengthen the robustness of deep learning models by reducing the noise introduced by these imperfections, as demonstrated by studies linking diagnostic accuracy reductions to the presence of artifacts [15]. Despite advancements, tools like HistoQC [16] face limitations in generalizability due to single-cohort training, underscoring the need for adaptable artifact detection across diverse datasets, staining protocols, and tissue types.

In this work, we address these gaps by proposing a robust artifact detection pipeline that integrates three complementary approaches: a deep learning-based approach (DLA) utilizing a fine-tuned ResNet50 architecture, a foundation model-based approach (FMA) leveraging a fine-tuned Unified Neural Image (UNI) model, and a knowledge-based approach (KBA) employing texture, color, and frequency metrics. Our pipeline not only detects a broader range of artifacts—including blood and tissue damage—but also provides enhanced adaptability and granularity compared to existing methods. The Adaptability is provided by the foundation model backbone, and the granularity is due to our focus on individual image artifacts.

The rest of this paper is organized as follows: Section II describes the datasets and methods used for artifact detection; Section III presents the experimental results; and Section IV discusses the implications of our findings and concludes the paper.

## II. Materials and Methods

In this study, we developed a multi-branch pipeline for artifact detection in whole slide images (WSIs), as summarized in Figure 2. The pipeline begins with data collection, where WSIs are gathered from multiple datasets representing diverse demographic and clinical characteristics. During the WSI processing phase, raw images are analyzed to extract embedded metadata, including micron-per-pixel (MPP) resolution, staining techniques (e.g., H&E, IHC), and magnification settings (e.g., 20X/40X), which are critical for assessing image quality and context. In the WSI processing phase, WSIs are segmented into smaller patches, followed by the application of the AI approach. The image patches are then evaluated in the patch scoring phase, where each patch is assessed for quality based on its representation of artifact-free target tissue using each of these approaches. Finally, the pipeline includes a visualization phase, providing insights into the distribution of artifacts. This step enables a comprehensive evaluation of artifact detection performance across entire WSIs, supporting both diagnostic and quality control objectives.

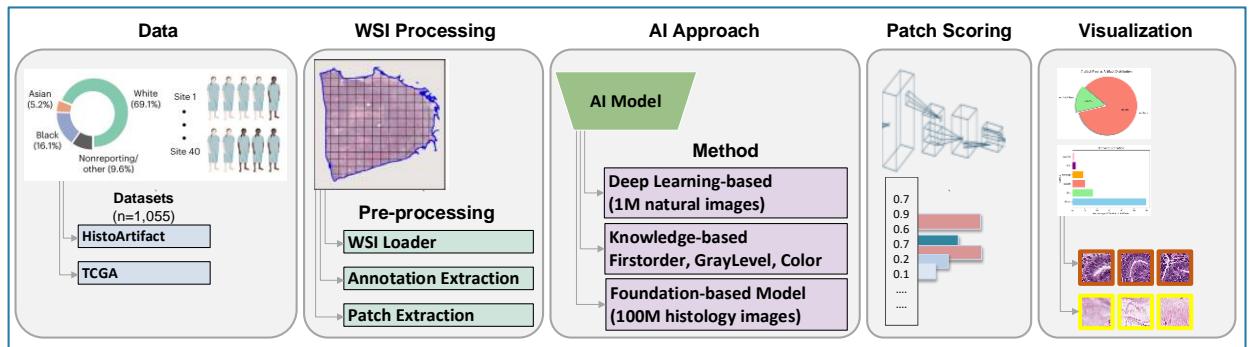

**Figure. 2** Overview of WSI artifact detection pipeline.

**Dataset**

The HistoArtifacts dataset [2] is a comprehensive, publicly available resource designed to support the development and evaluation of artifact detection algorithms in histopathology WSIs. It includes annotated image patches extracted from WSIs, encompassing six common artifact types: air bubbles, out-of-focus regions, blood, tissue folds, tissue damage, and marker artifacts. Each patch is annotated at the pixel level. This dataset is representative of diverse staining protocols, tissue types, and imaging conditions, which provides a robust foundation for training and validating computational pathology models. It features WSIs scanned using multiple platforms, including Philips, Leica Aperio, Hologic, Hamamatsu, and 3D Histech, and includes widely used staining protocols, such as Hematoxylin and Eosin (H&E) and Formalin-Fixed methods. This device and protocol diversity ensures the generalizability of algorithms across different imaging systems, enhancing the validation of the proposed pipeline. The dataset comprises 55 WSIs and a total of 52,780 image patches, cropped to 224x224 pixels, which were divided into three subsets: training, tuning, and test sets. The training set consists of 34,408 patches from 35 WSIs used for model development, while the tuning set, with 8,114 patches, gathered from 10 WSIs, is used for hyperparameter optimization. The hold out test set, comprising 10,258 patches, extracted from 10 independent WSIs, is reserved for evaluating model performance. These subsets are further categorized by artifact type: artifact-free, blur, tissue folds, tissue damage artifacts, and bubble artifacts. This categorization ensures comprehensive testing of models on a wide range of artifact types. The dataset's detailed breakdown of WSIs and patches, along with their allocation across subsets, is presented in Table 1. In Figure 5, the ratio of artifact-free vs all artifacts is shown in 5(a) and in 5 (b) the distribution of the artifacts themselves can be seen. Artifact-free comprises about 15% of the dataset, while the artifacts represent around 85%. Among the artifacts themselves, blood and blur are most represented, while marker and are least represented. The datasets used in this study are publicly available and can be accessed through the respective data repositories. Specifically, the TCG and HistoArtifact datasets used in this research are available through The Cancer Genome Atlas (TCGA) portal and Zotero website. Additional data processing steps and any custom code or models used in this study is available on the project GitHub page (https://github.com/DIDSR/HistoART).

**Table 1** A breakdown of the number of WSIs and patches used for model development, tunning, and testing.

| Dataset | Scanner | Staining Protocol | All | | Training | | Tuning | | Testing | |
|---|---|---|---|---|---|---|---|---|---|---|
| | | | #Patches (#WSIs) | Category | Patches | Category | Patches | Category | Patches | Category |
| TCGA/ Histo Artifact | Hamamatsu, Hologic, Aperio Leica AT2, Philips | H&E, Formalin-Fixed | 52780 (55) | Artifact Free 7805 | 34408 | 5249 | 8114 | 1591 | 10258 | 965 |
| | | | | Out of Focus 7552 | | 5661 | | 754 | | 1137 |
| | | | | Tissue Fold 1243 | | 998 | | 114 | | 131 |
| | | | | Marker 851 | | 681 | | 0 | | 170 |
| | | | | Air Bubble 4520 | | 2499 | | 1175 | | 846 |
| | | | | Blood 26887 | | 16743 | | 4148 | | 5996 |
| | | | | Tissue Damage 3922 | | 2577 | | 332 | | 1013 |

**Development**

In this study, we propose three approaches for artifact detection and reporting in WSIs 1) a foundation-based model approach, 2) a deep learning CNN-based approach, and 3) a knowledge-based approach. Each approach contributed unique strengths to the pipeline, providing a comprehensive representation of the image data for detecting artifacts in WSIs.

**Foundation-Based Model Approach (FMA):** For the FMA, we finetuned UNI, a foundational vision model specifically designed for histopathology tasks. UNI is based on the Vision Transformer (ViT-16) architecture and was pre-trained on the MASS-100k dataset, which consists of 100,000 WSIs spanning 20 organ types and includes over 100 million image patches. This large-scale pretraining enabled UNI to learn rich, generalizable feature representations that can be adapted to specific tasks. The training process for UNI involved taking WSI patches, generating multiple views, and extracting both masked crops and local crops from each view. These crops were then used to train the model for classification and segmentation tasks, including the detection of various artifacts. For our study, we focused exclusively on UNI's classification capabilities. To fine-tune UNI for artifact detection, we used the HistoArtifact training and tuning datasets. There is no overlap between the HistoArtifact dataset and UNI's pretraining data, ensuring unbiased evaluation. For each classification task, we selected relevant subsets of the data, either comparing artifact-free patches to a specific artifact type or treating all artifacts as a single class. During finetuning, we froze all but the last two encoder blocks of UNI's transformer architecture to accelerate training and reduce the risk of overfitting. This approach allowed us to adapt the model efficiently while preserving the general features learned during pretraining. After freezing the earlier layers, we finetune the model for ten epochs, using the Adam optimizer, batch size of 16, and a learning rate of 1e-4. This process was run on a workstation with an NVIDIA RTX 3060, AMD 5950x CPU, and 80GB of DRAM.

**Deep Learning-Based Approach (DLA):** The DLA utilized convolutional neural networks (CNNs) to automatically learn and extract complex features from histopathological images. Specifically, we employed a ResNet50 architecture, pre-trained on ImageNet [17], and finetuned it for artifact detection tasks using the HistoArtifact training dataset. Similarly to the FMA approach, we froze all convolutional layers except the last one in order to accelerate finetuning. This training process for this model also used the Adam optimizer, batch size of 16, ten epochs, and a learning rate of 1e-4. We used the tuning set to tune these hyperparameters. The hardware used is the same as for the FMA. CNNs excel at capturing hierarchical feature representations, enabling the detection of subtle patterns and relationships within the images. This approach was particularly effective at identifying intricate textures and morphological details that are difficult to encode manually.

**Knowledge-Based Approach (KBA): Feature Selection and Manifold Analysis:** While the FMA and DLA models demonstrates strong performance across various artifact types, gaining deeper insights into the specific features that contribute most to class differentiation, model explainability is important. To address this, we employed a knowledge-based approach (KBA) that relies on handcrafted feature extraction and feature space visualization to evaluate and interpret artifact detection in WSIs.

The KBA focused on selecting domain-specific handcrafted features that capture critical characteristics of WSIs. Features were grouped into three main categories: Texture Features: Metrics derived from gray-level co-occurrence matrix (GLCM) [18], fractal dimension texture analysis [19], and local binary patterns (LBP) [20] to capture spatial relationships and structural patterns in tissue. GLCM accounts for 44 total features, fractal dimension 4 features, and LBP provide 30 features. Color Features: Features based on pixel intensity distributions, hue-saturation-value (HSV) histograms, and stain absorption are designed to represent staining intensity and color variations [21]. Color variations and average HSV contain 3 features each, while co-matrices use 26 features and entropy 9 features. Frequency Domain Features: Fourier transform-based metrics to quantify structural periodicity and noise characteristics are particularly useful for identifying artifacts such as blurring or tissue folds [22]. The Fourier power spectrum accounts for 15 total features. These handcrafted features provided interpretable insights into artifact presence by isolating patterns linked to specific artifact types. While these features may lack the generalization capabilities of the FMA and DLA models, they complement them by offering explainable features.

To better understand the discriminative power of these features, we employed dimensionality reduction and visualization techniques, including t-SNE [23], Principal Component Analysis (PCA) [24], and Uniform manifold approximation (UMAP) [25], to explore the structure of the feature space using the tuning data. This analysis aimed to evaluate how effectively different feature sets separate artifact-free images from those with artifacts. Artifact-Free vs. Blur (Out-of-Focus) Artifacts: Figure 3a illustrates that GLCM and LBP features exhibited the clearest separability between classes, forming well-defined clusters with minimal overlap, particularly for the out of focus class. In contrast, Color and Fourier features provided weaker differentiation. Artifact-Free vs. Tissue Fold Artifacts: Figure 3b shows a similar trend, where GLCM and LBP demonstrated the strongest separability. However, the distinction was less pronounced, likely due to the complexity of tissue folds and imbalanced dataset composition, with only 1,200 tissue fold images compared to 7,800 artifact-free and 7,500 blur images. When comparing artifact-free images against all artifact types (Figure 3c), the inclusion of less discriminative features (e.g., Fourier and Color) reduced the separation between classes. This observation highlights the importance of feature selection in improving classification performance. These findings underscore that a subset of highly discriminative features, such as GLCM and LBP, enhance the ability to distinguish artifacts effectively. In Figure 3(d) we point out the areas which were difficult for the models to classify for out of focus and tissue fold, such as samples that reside in the boundary areas between classes. While this overlap between classes remained, especially with mixed artifact types, the results demonstrated that a focused feature set improves model efficiency and accuracy. In the following section, we will show how using a subset of the best performing features will allow us to build a Support Vector Machine (SVM) [26] model that performs close to or better than state-of-the art deep learning and foundational models.

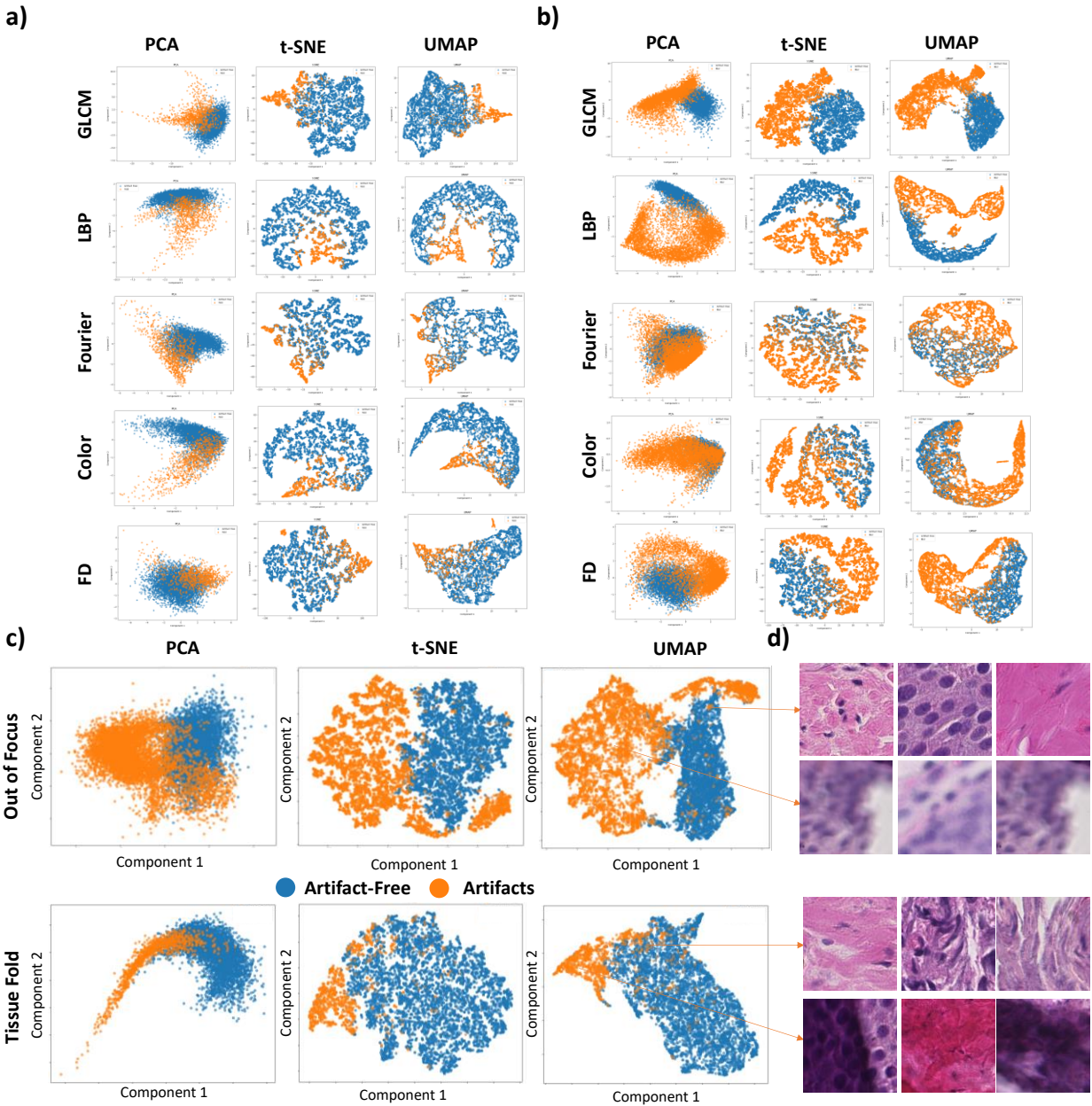

**Figure. 3** In (a) tissue fold and (b) out of focus, the t-SNE, PCA, and UMAP derived from the tuning data are shown above for each of the listed features: GLCM, LBP, Fourier, color, and fractal dimension. The first two components for each of the features are used for the plot. As we can see, GLCM and LPB provide the neatest separation for the out-of-focus class, but less so for tissue fold. In (c) we use all features in the t-SNE, PCA, and UMAP models. Note that the addition of the less performant features inhibits the separation of the classes, as the overlap increases in comparison to the GLCM and LPB plots for out-of-focus. In the patch images shown in (d), samples that were difficult for the models to classify correctly are shown in the first and third rows. Images with a ground truth of artifact free are marked as out of focus or tissue fold, respectively, in these rows. The second and fourth rows are easier to classify as they have clear distinct features of the artifacts. The arrows from the component analyses illustrate their position in feature space.

Each approach— FMA, DLA, and KBA—was independently developed and evaluated to assess its unique contributions to artifact detection. The KBA provided interpretability and domain-specific insights, which complement the intricate pattern recognition capabilities of deep learning and the generalization strengths of foundation models. Together, these methods enable a comprehensive evaluation framework, with handcrafted features offering an essential layer of analysis for understanding artifact-specific patterns.

The primary performance figure of merit for assessing our model in the task of predicting artifacts in WSIs is the area under the receiver operating characteristic (ROC) curve (AUC) [27]. ROC curves are plots of sensitivity against 1-specificity, where sensitivity is defined as the number of predicted artifact-free images divided by the total number of artifact-free images (also known as true positive rate (TPR)), and the specificity (1-false positive rate (FPR)) is defined as the number of predicted artifacts divided by the total number of artifacts (i.e., artifacts are defined as "positive" and artifact-free are defined as "negative; see Equation 1). We used the DeLong et al. [28] method for calculating the uncertainty and confidence intervals of all the AUC estimates. We also examined secondary performance metrics such as precision, recall, F1-score, and accuracy (Equation 2) for the algorithm predictions dichotomized at a cut-off value of 0.5. The value of this threshold is often the default choice of 0.5 [29].

$$\text{TPR} = \text{Sensitivity} = \text{Recall} = \frac{TP}{TP+FN}, \text{FPR} = 1 - \text{specificity} = \frac{FP}{TN+FP}, \text{Precision} = \frac{TP}{TP+FP}, \tag{1}$$

$$F1 = 2 \cdot \frac{\text{Precision} \cdot \text{Recall}}{\text{Precision}+\text{Recall}} = \frac{2TP}{2TP+FP+FN}, \text{Accuracy} = \frac{TP+TN}{TP+TN+FN+FP}, \tag{2}$$

## III. Results

The results of the binary classification task of artifact-free vs all artifacts are shown in Figure 4 (a). As expected, the FMA outperforms the DLA and KBA, with tighter confidence intervals. However, all three models achieve high AUC scores. The analyses for one vs. all artifact multiclassification are shown in Figure 4 (b), (c), and (d), for the DLA, FMA, and KBA respectively. The FMA achieves near perfect results for all artifacts, including those with smaller sample sizes in the training set. The DLA model and KBA also perform well on the larger sample sizes, such as out-of-focus and tissue fold, but struggle with true negatives for the smaller marker and tissue damage.

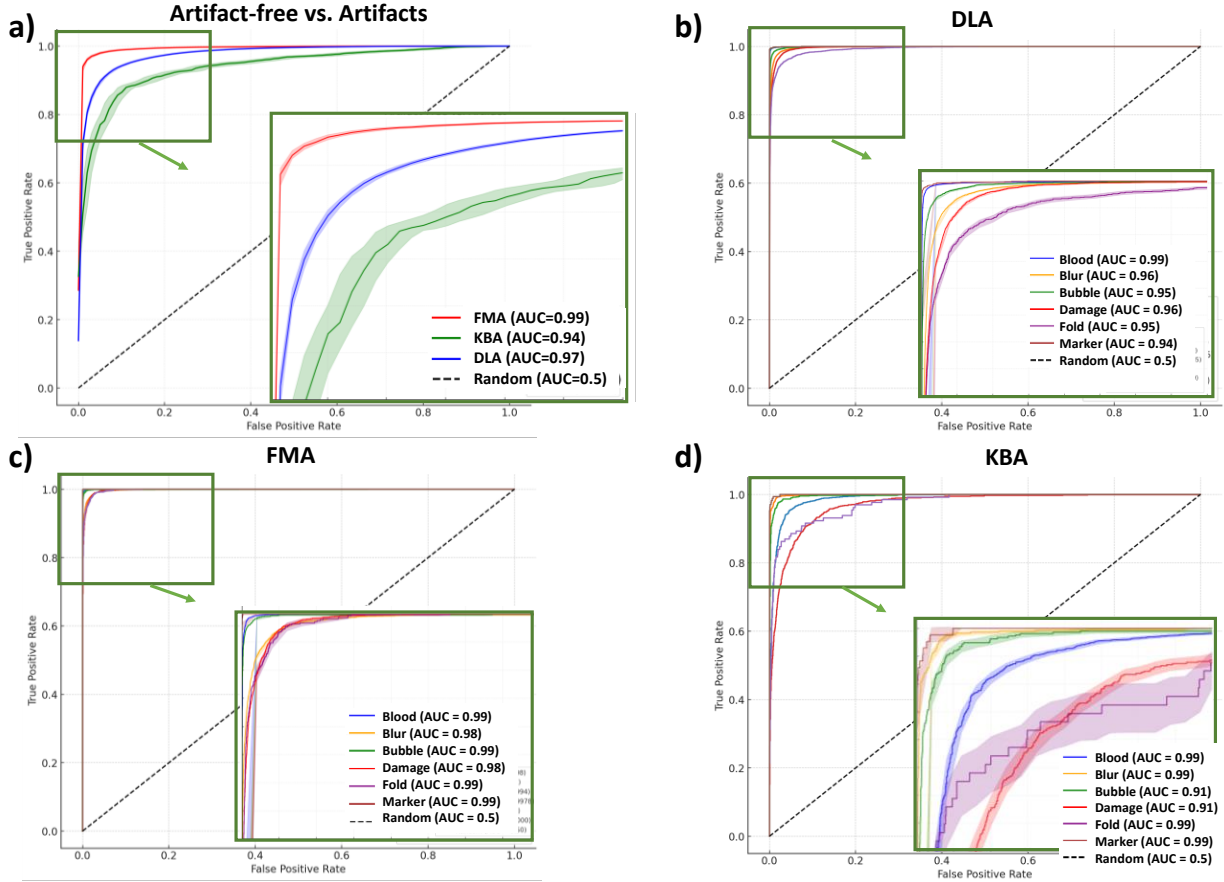

**Figure. 4** Test results of our WSI artifact detection method. a) ROC curves of FMA, DLA, and KBA methods for artifact vs. non-artifact detection. B. ROC curves of DLA for individual artifact types when compared to Artifact-free patches, c) ROC curves of FMA for individual artifact types when compared to artifact-free patches. d) ROC curves of KBA for individual artifact types when compared to artifact-free patches.

The results are presented in Table 2. The FMA achieved an AUC of 0.995 with 95% CIs [0.994, 0.995], which is higher than the AUC of the DLA, which was 0.977 [0.977, 0.978], and KBA which was 0.940 with 95% CIs [0.933 0.946]. While the FMA performed better compared to the DLA, and KBA, the difference was not significant. Nevertheless, our findings suggest that the FMA could be more effective approach for detecting artifacts.

**Table 2** Performance of DLA, FMA, and KBA models for artifact detection task. Note: a score cutoff of 0.5 was used for estimating the precision, recall, F1 and accuracy values reported in the table.

| Model | Precision [CI] | Recall [CI] | F1-Score [CI] | Accuracy [CI] | AUC [CI] |
|---|---|---|---|---|---|
| DLA | 0.958 [0.957 0.960] | 0.980 [0.979 0.9] | 0.97 [0.97 0.97] | 0.947 [0.946 0.948] | 0.977 [0.977 0.978] |
| FMA | **0.990** [0.989 0.991] | **0.982** [0.981 0.983] | **0.99** [0.99 0.99] | **0.976** [0.975 0.977] | **0.995** [0.994 0.995] |
| KBA | 0.969 [0.965 0.973] | 0.940 [0.935 0.945] | 0.954 [0.94 0.95] | 0.919 [0.913 0.924] | 0.940 [0.933 0.946] |
| Kanwal et al. (2024) | | | 0.968 | 0.963 | 0.96 |

Table 3 provides a comprehensive comparison of the performance of the three artifact detection methods—DLA, FMA, and KBA—for detecting the six common artifact types. The results are also compared with previously reported performance metrics from Kanwal et al. (2024), where applicable. Kanwal et al. (2024) uses the same test set we use in this study. The AUC values across our three models and six artifact types highlight the effectiveness of each model in distinguishing between classes. For tissue fold, the FMA model achieves the highest AUC (0.998), followed closely by KBA (0.991) and DLA (0.953). In the out of focus condition, all models perform exceptionally well, with KBA achieving nearly perfect discrimination (0.9997), while FMA and DLA also maintain strong AUC values of 0.998 and 0.963, respectively. Marker classification sees KBA and FMA achieving AUC values (0.9998 and 0.9999), whereas DLA lags behind at 0.946. Air bubble detection results in FMA leading at 0.995, followed by DLA (0.959) and KBA (0.910). For tissue damage, FMA again leads with an AUC of 0.989, slightly outperforming DLA (0.969), and both outperforming KBA which scores at 0.917. Finally, in blood artifact classification, all methods perform exceptionally well, with FMA leading at 0.9996, followed closely by KBA (0.993) and DLA (0.997). These results indicate that while all models demonstrate strong classification abilities, FMA and KBA generally outperform DLA.

The FMA consistently outperforms the other methods across most artifact types, demonstrating its ability to balance precision, recall, and overall accuracy. Its high AUROC values indicate strong generalizability across diverse artifact types and imaging conditions (e.g., scanners, stains). We believe this is due to the massive pretraining set used to build the FMA, allowing it to excel in constrained downstream tasks.
KBA offers competitive performance for specific artifact types, such as out-of-focus regions and marker artifacts, where handcrafted features may have distinct advantages. However, its performance tends to decline for more complex artifact types like tissue damage and air bubbles, highlighting the limitations of feature engineering alone.
DLA exhibits good performance, often achieving good precision but struggling with recall for several artifact types. This suggests it may miss subtle artifacts and highlights the need for further optimization.
Compared to Kanwal et al., the proposed FMA model and KBA consistently achieve equal or superior performance metrics, particularly in recall and AUROC. The table clearly demonstrates the superiority of the FMA in achieving high accuracy and robustness across all artifact types. It underscores the importance of leveraging a FMA for reliable artifact detection, while also validating the complementary but distinct role of handcrafted features in specific contexts. These findings are promising and provide a step toward enhanced artifact detection pipelines that provide both interpretability and state-of-the-art performance for digital pathology workflows.

**Table 3** Performance of DLA, FMA, and KBA models for each artifact type. Note: a score cutoff of 0.5 was used for estimating the precision, recall, F1 and accuracy values reported in the table.

| | Method | Precision [CI] | Recall [CI] | Accuracy [CI] | AUC [CI] |
|---|---|---|---|---|---|
| **Tissue Fold** | DLA | 0.887 [0.873 0.901] | 0.518 [0.501 0.535] | 0.922 [0.919 0.926] | 0.953 [0.950 0.957] |
| | FMA | **0.941** [0.927 0.954] | **0.956** [0.944 0.68] | **0.995** [0.983 0.988] | **0.998** [0.997 0.998] |
| | KBA | 0.810 [0.748 0.869] | 1.000 [1.000 1.000] | 0.972 [0.962 0.982] | 0.991 [0.986 0.996] |
| | Kanwal et al. (2024) | | | 0.9328 | 0.930 |
| **Out of Focus** | DLA | 0.974 [0.973 0.976] | 0.95 [0.951 0.956] | 0.963 [0.961 0.964] | 0.963 [0.962 0.964] |
| | FMA | 0.986 | 0.986 | 0.986 | 0.998 |
|  |  | [0.983 0.989] | [0.983 0.989] | [0.984 0.988] | [0.998 0.999] |
|  | KBA | **0.996** [0.993 1.000] | **0.996** [0.992 0.999] | **0.996** [0.993 0.999] | **0.9997** [0.9994 1.000] |
|  | Kanwal et al. (2024) |  |  | 0.995 | 0.995 |
| Marker | DLA | 0.880 [0.846 0.914] | 0.302 [0.274 0.330] | 0.965 [0.963 0.967] | 0.946 [0.939 0.953] |
|  | FMA | 0.980 [0.965 0.995] | 0.985 [0.972 0.998] | **0.998** [0.997 0.999] | **0.9999** [0.9998 1.0] |
|  | KBA | **0.994** [0.983 1.000] | **0.990** [0.972 1.000] | 0.996 [0.993 1.000] | 0.9998 [0.9994 1.0] |
|  | Kanwal et al. (2024) | N/A | N/A | N/A | N/A |
| Air Bubble | DLA | 0.938 [0.934 0.944] | 0.693 [0.684 0.702] | 0.884 [0.881 0.888] | 0.959 [0.956 0.961] |
|  | FMA | **0.968** [0.962 0.974] | **0.933** [0.925 0.942] | **0.968** [0.965 0.972] | **0.995** [0.994 0.996] |
|  | KBA | 0.942 [0.921 0.963] | 0.517 [0.483 0.550] | 0.760 [0.740 0.780] | 0.910 [0.893 0.920] |
|  | Kanwal et al. (2024) | N/A | N/A | N/A | N/A |
| Tissue Damage | DLA | 0.944 [0.939 0.950] | 0.781 [0.773 0.790] | 0.910 [0.909 0.912] | 0.969 [0.967 0.971] |
|  | FMA | **0.970** [0.964 0.976] | **0.869** [0.857 0.880] | **0.945** [0.941 0.950] | **0.989** [0.988 0.991] |
|  | KBA | 0.881 [0.860 0.903] | 0.777 [0.751 0.803] | 0.832 [0.816 0.849] | 0.917 [0.905 0.930] |
|  | Kanwal et al. (2024) | N/A | N/A | N/A | N/A |
| Blood | DLA | 0.978 [0.977 0.980] | 0.995 [0.994 0.995] | 0.979 [0.978 0.980] | 0.997 [0.997 0.997] |
|  | FMA | **0.997** [0.996 0.998] | **0.997** [0.996 0.998] | **0.996** [0.995 0.996] | **0.9996** [0.9994 0.9999] |
|  | KBA | 0.990 [0.987 0.992] | 0.971 [0.970 0.975] | 0.966 [0.962 0.970] | 0.993 [0.990 0.994] |
|  | Kanwal et al. (2024) | N/A | N/A | N/A | N/A |

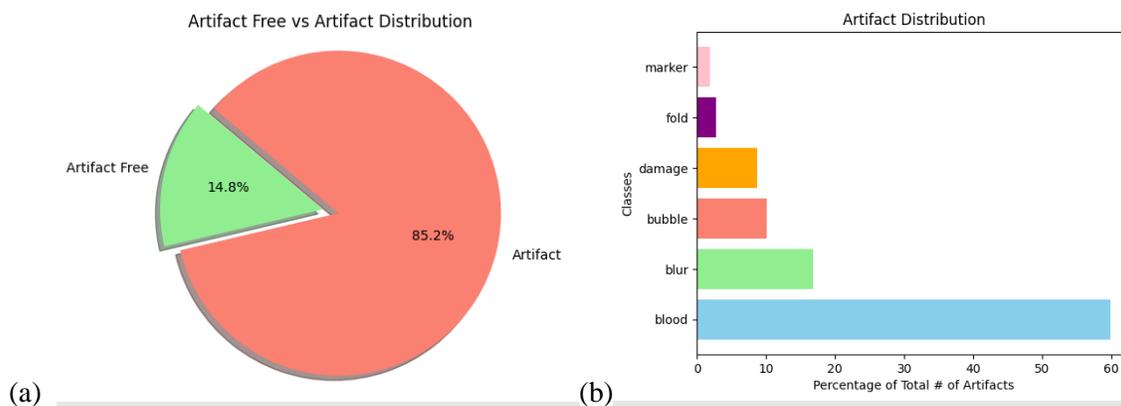

**Figure 5.** In (a) the percentage of artifacts and artifact-free samples are shown as a percentage of the total number of samples in the dataset. In (b) the distribution of artifacts in the dataset are shown as a percentage of the total number of artifacts. The blood class dominates, followed by blur, bubble, damage, fold, and marker.

## IV. Discussion

In this study, we developed and evaluated a multi branch pipeline for artifact detection in WSIs, with the aim to improve the reliability of CPATH systems by addressing common artifacts such as out of focus blur, tissue folds, air bubbles, and marker artifacts. Our findings demonstrate that FMA and KBA outperform the DLA method.

The FMA, which leverages UNI pre-trained models, outperformed both the DLA and KBA approaches for most types of artifacts. These foundation models, pretrained on large and diverse datasets, offer significant advantages in terms of transferability and adaptability to smaller, domain-specific datasets like those in medical imaging. Fine-tuning such models for artifact detection demonstrated their ability to generalize across various artifact types, even with limited annotated data. As shown in a Comprehensive Evaluation of Histopathology models [30], foundation models such as UNI, which are pretrained on massive histopathology datasets, outperform their deep learning-based counterparts. Given that the authors compared the performance of FMA and DLA on a dataset previously unseen by the models, this suggests better performance by pre-trained foundation models in general.

While the DLA exhibited reasonable performance, the "black-box" nature of deep learning and foundation models means they lack interpretability and explainability. On the other hand, the KBA, which combines texture, color, and frequency domain features, provides a somewhat more interpretable and transparent solution, allowing for greater insight into the decision-making process. Therefore, KBA could be more suitable for environments where interpretability is important. However, handcrafted features can struggle to capture the full complexity of WSIs, especially when artifacts exhibit irregular or subtle patterns that are difficult to quantify manually. Overall, our results indicate that foundation models offer superior performance, while handcrafted features remain a valuable option for ensuring transparency and interpretability. Deep learning models, though highly effective, face challenges related to resource requirements and lack of transparency, limiting their application in certain clinical contexts.

The proposed artifact detection approach has the potential to improve the clinical workflow in digital pathology by enhancing the accuracy and reliability of diagnostic outcomes. By minimizing the impact of artifacts on computational pathology systems, this method could potentially help ensure that critical diagnostic features in WSIs are preserved and accurately analyzed. This is particularly important in tasks such as tumor detection, grading, and classification, where even subtle artifacts can lead to misdiagnoses [31]. Ultimately, this tool may enhance patient care by ensuring more consistent, high-quality diagnostic data and fostering confidence in computational pathology systems.

Accurate artifact detection has additional valuable use cases beyond improving clinical diagnostics. One important application is in flagging issues with current datasets before they are used for training or validating AI models. Current datasets used in computational pathology may include images with varying levels of quality, and artifacts within these images can introduce noise that reduces the effectiveness of AI training. By detecting and flagging such artifacts, this tool can help curate cleaner datasets. Moreover, artifact detection can play a pivotal role in quality assurance workflows for WSI scanners and laboratory processes. Regularly monitoring and identifying artifact-prone slides can highlight issues with scanning devices, staining protocols, or sample preparation methods, allowing for timely interventions to maintain data integrity.

In conclusion, our study highlights the strengths of deep learning models, foundation models, and handcrafted feature-based approaches in advancing artifact detection in WSIs. By enhancing artifact detection, these methods have the potential to improve the reliability and robustness of computational pathology systems, ultimately leading to better diagnostic accuracy. The methods from this work are available for download at https://github.com/DIDSR/HistoART.

## Author Contributions

Seyed Kahaki: Conceptualized and designed the study, drafted the original manuscript, developed the code, data preparation, and performed analyses. Alexander Webber: Reviewed and revised the manuscript, developed the code, data preparation, trained and tested models, and conducted formal analyses. Ghada Zamzmi: Drafting the original manuscript, contribute to the study design, and reviewed and revised the manuscript. Adarsh Subbaswamy: Reviewed and revised the manuscript. Rucha Deshpande: Reviewed and revised the manuscript. Aldo Badano: Supervised the project, and reviewed and revised the manuscript.

## Competing interests

The authors declare no competing interests.